\documentclass[conference]{IEEEtran}
\IEEEoverridecommandlockouts
\usepackage{cite}
\usepackage{amsmath,amssymb,amsfonts}
\usepackage{graphicx}
\usepackage{textcomp}
\usepackage{xcolor}

\usepackage{algorithm}
\usepackage[noend]{algorithmic}

\usepackage{booktabs}
\usepackage{multirow} 
\usepackage{array}
\usepackage{url}
\usepackage{cite}

\usepackage{hyperref}
\hypersetup{
    colorlinks=true,
    linkcolor=red,
    citecolor=green,
    filecolor=magenta,
    urlcolor=cyan
}

\def\BibTeX{{\rm B\kern-.05em{\sc i\kern-.025em b}\kern-.08em
    T\kern-.1667em\lower.7ex\hbox{E}\kern-.125emX}}
\begin{document}

\renewcommand{\arraystretch}{0.5}
\setlength{\aboverulesep}{1.25pt} 
\setlength{\belowrulesep}{1.25pt} 
\setlength{\textfloatsep}{5pt}

\title{
PGD-Imp: Rethinking and Unleashing Potential of Classic PGD with Dual Strategies for Imperceptible Adversarial Attacks
}

\newcommand{\linebreakand}{%
  \end{@IEEEauthorhalign}
  \hfill\mbox{}\par
  \mbox{}\hfill\begin{@IEEEauthorhalign}
}

\author{
\IEEEauthorblockN{Jin Li$^{1}$, Zitong Yu$^{2}$, Ziqiang He$^{1}$, Z. Jane Wang$^{3}$, Xiangui Kang$^{1*}$\thanks{*\,Xiangui Kang is the corresponding author. 
This work has been accepted to the proc. of 50th IEEE ICASSP. Copyright may be transferred without notice, after which this version may no longer be accessible.
}}
\IEEEauthorblockA{$^{1}$\textit{Guangdong Key Lab of Information Security}, \\ \textit{School of Computer Science and Engineering, Sun Yat-Sen University} \\ $^{2}$\textit{School of Computing and Information Technology, Great Bay University} \\$^{3}$\textit{Electrical and Computer Engineering Dept, University of British Columbia}}
}
\maketitle

\begin{abstract}
Imperceptible adversarial attacks have recently attracted increasing research interests. Existing methods typically incorporate external modules or loss terms other than a simple $l_p$-norm into the attack process to achieve imperceptibility, while we argue that such additional designs may not be necessary. In this paper, we rethink the essence of imperceptible attacks and propose two simple yet effective strategies to unleash the potential of PGD, the common and classical attack, for imperceptibility from an optimization perspective. Specifically, the Dynamic Step Size is introduced to find the optimal solution with minimal attack cost towards the decision boundary of the attacked model, and the Adaptive Early Stop strategy is adopted to reduce the redundant strength of adversarial perturbations to the minimum level. The proposed PGD-\textit{Imperceptible} (PGD-Imp) attack achieves state-of-the-art results in imperceptible adversarial attacks for both untargeted and targeted scenarios. When performing untargeted attacks against ResNet-50, PGD-Imp attains 100$\%$ (+0.3$\%$) ASR, 0.89 (-1.76) $l_2$ distance, and 52.93 (+9.2) PSNR with 57s (-371s) running time, significantly outperforming existing methods. 
\end{abstract}


\section{Introduction}
Adversarial attacks have revealed the vulnerability of deep learning models \cite{42503, goodfellow2014explaining, yuan2019adversarial}. 
While many studies investigated the attack performance and transferability \cite{madry2018towards, dong2018boosting, jin2023multi, fan2023enhance, feng2023dynamic} under $l_p$-norm distances, recently another line of increasing studies focused on enhancing the imperceptibility of attacks \cite{carlini2017towards,luo2018towards,zhao2020towards,laidlaw2021perceptual,duan2021advdrop,jia2022exploring, luo2022frequency, chen2023imperceptible}, since the simple $l_p$-norm restrictions were shown inadequate in deceiving the Human Visual System (HVS) \cite{sharif2018suitability}.

Existing restricted imperceptible adversarial attacks typically incorporate perception-related modules or losses into the attack process. The aim is to leverage the perceptual characteristics of the HVS to impose more constraints and limit the adversarial perturbations
thereby enhancing imperceptibility. For example, 
PerC-AL \cite{laidlaw2021perceptual} introduced a color perceptual distance and a corresponding loss function.
AdvDrop \cite{duan2021advdrop} employed the Discrete Cosine Transform (DCT) to discard image details in the frequency domain.
SSAH \cite{luo2022frequency} proposed to optimize in feature space with constraint based on Discrete Wavelet Transform (DWT), and AdvINN\cite{chen2023imperceptible} also used the DWT combined with invertible neural networks to manipulate the category-related information for targeted attacks.

\begin{figure}[t]
\centering
\includegraphics[width=0.8\columnwidth]{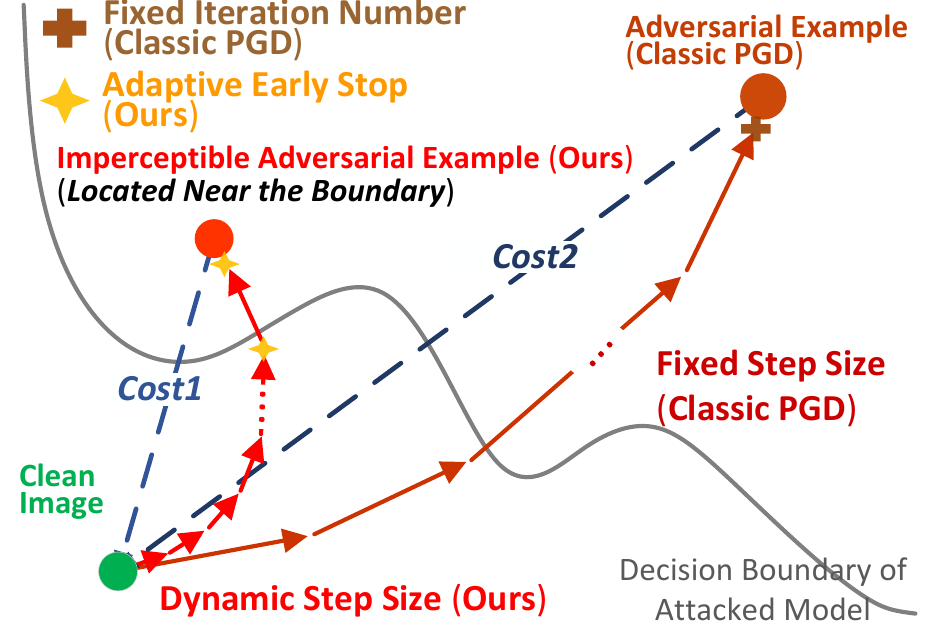}
\vspace{-0.4em}
\caption{Illustration of our key idea and the proposed method. 
The essence of imperceptible adversarial attacks lies in pushing the image across the decision boundary of the attacked model with minimal cost. To fulfill this, we propose two simple yet ingenious strategies, Dynamic Step Size and Adaptive Early Stop, to unleash the potential of classic PGD for imperceptibility.
}
\label{fig:1}
\end{figure}

Despite the above progress, for the first time, we realize that such additional complex designs specifically proposed for imperceptibility may not be necessary. Rather than relying on externally hiding or constraining adversarial perturbations, we argue that the essence of imperceptible adversarial attacks lies in pushing the image across the decision boundary of the attacked model with minimal cost. In other words, if an attack method can successfully perform with minimal perturbation strength, the attack is inherently imperceptible, eliminating the need for further constraints or conditions. Consequently, by viewing imperceptible attacks as an optimization problem concerning the model decision boundary, we propose the PGD-\textit{Imperceptible} (PGD-Imp) attack with two simple yet ingenious strategies to unleash potential of the classical Projected Gradient Descent (PGD) \cite{madry2018towards} for imperceptibility, considering PGD is one of the most foundational algorithm in adversarial attacks and has been proven to be excellent and flexible to solve the optimization problem of adversarial attacks \cite{dong2019evading, lin2019nesterov, zhao2021success, wang2021enhancing, zhang2023improving}. 

The key idea of our method is illustrated in Fig.~\ref{fig:1}. Firstly, we propose the Dynamic Step Size strategy to replace the fixed step size in the original PGD inspired by the widely adopted learning rate schedules in the field of optimization \cite{kingma2014adam, loshchilov2016sgdr, loshchilov2017decoupled, ruder2016overview}.
More concretely, instead of using fixed integer step size with projection operation to meet the perturbation budget, we use smaller and dynamic step sizes to allocate the total budget to all iterations, refining the optimization process.
This strategy, which is overlooked in previous attacks, effectively enables the optimization process to converge toward an optimum, reaching the decision boundary with a lower cost. Secondly, considering that desired adversarial examples should just slightly cross the decision boundary and remain close to minimize redundant perturbations, we employ an Adaptive Early Stop strategy in the later steps of attack,
which halts the optimization as soon as the current adversarial example successfully attacks. In line with the early stopping in model training to prevent the models from overfitting to the training set \cite{prechelt2002early, liu2008optimized, bai2021understanding}, this strategy for attacks also prevents the optimized adversarial examples in overfitting to redundantly cross the decision boundary, thereby keeping the attack cost to a minimum level.

Extensive experiments are conducted to demonstrate the effectiveness and superiority of the proposed PGD-Imp. Equipped with the above dual strategies, our PGD-Imp simultaneously achieves high imperceptibility and attack efficacy in both untargeted and targeted scenarios at a even shorter running time as an improved attack pipeline developed from a novel perspective of optimization. 

\section{Methodology}
\subsection{Preliminary}

For the image classification task, given a benign image $\boldsymbol{x}$ with the ground-truth label $y_{gt}$ and a well-trained classifier $f(\cdot)$, untargeted attacks aim to craft adversarial example $\boldsymbol{x}_{adv} = \boldsymbol{x} + \boldsymbol{\delta}$ that can mislead the classifier to output wrong prediction $f(\boldsymbol{x}_{adv}) \neq y_{gt}$, where $\boldsymbol{\delta}$ is the injected adversarial perturbation. 
When all information about the attacked model is transparent to attacker (i.e., white-box attacks), this process 
can be considered as an optimization problem as:
\begin{equation}
\mathop{\arg\max}_{\boldsymbol{\delta}} \ \mathcal L(f(\boldsymbol{x} + \boldsymbol{\delta}), y_{gt}), \quad  s.t. \|\boldsymbol{\delta}\|_\infty \leq \epsilon,
\end{equation}
where $\mathcal L$ is the loss function (e.g., Cross-Entropy $\mathcal L_{\text{CE}}$), and $\|\boldsymbol{\delta}\|_\infty \leq \epsilon$ is a basic restriction to limit the maximum perturbation intensity of $\boldsymbol{\delta}$ under the budget $\epsilon$.
Similarly, the form of targeted attacks can be easily obtained attacks by replacing $y_{gt}$ with $y_{tar}$ and argmax with argmin for misleading the classifier to output the specified category $f(\boldsymbol{x}_{adv}) = y_{tar}$. 

As a milestone in adversarial attacks, PGD \cite{madry2018towards} is a powerful solution to this optimization problem, which iteratively updates the image using the current gradient direction, and then projects the obtained solution back into the limited range. PGD for untargeted attack at step $t$ can be denoted as:
\begin{equation}
\boldsymbol{x}_{t+1} = \Pi_{x, \epsilon}\{\boldsymbol{x}_t + \alpha\cdot\text{sign}(\nabla_{\boldsymbol{x}_t}\mathcal L_{\text{CE}}(f(\boldsymbol{x}_{t}), y_{gt}))\},
\end{equation}
where $\Pi$ is the projection operation and $\alpha$ is the step size.

\begin{algorithm}[t]
    \footnotesize
    \caption{\textbf{Proposed PGD-Imp}}
    \label{alg1}
    \begin{algorithmic}[1] 
        \STATE \textbf{Input}: attack type (untargeted or targeted), image $\boldsymbol{x}$ with label $y_{gt}$ (or target label $y_{tar}$), attacked model $f(\cdot)$, perturbation budget $\epsilon$, step $T$;
        \STATE Calculate linear coefficient sequence $\eta_{1:T} \in {(0, 1]}^T$, and then calculate $\beta = {\epsilon} / {\sum_{t=1}^T \eta_t}$ in Eq.~\eqref{eq3};
        \STATE No random initialization, start with $\boldsymbol{x}_{1} = \boldsymbol{x}$;
        \FOR{$t=1$ to $T$}
            \IF {\textbf{untargeted attack}}
                \STATE $\boldsymbol{x}_{t+1} = \boldsymbol{x}_t + \eta_t\cdot\beta\cdot\text{sign}(\nabla_{\boldsymbol{x}_t}\mathcal L_{\text{CE}}(f(\boldsymbol{x}_{t}), y_{gt}))$ as Eq.~\eqref{eq4};
            \ELSIF{\textbf{targeted attack}}
                \STATE $\boldsymbol{x}_{t+1} = \boldsymbol{x}_t - \eta_t\cdot\beta\cdot\text{sign}(\nabla_{\boldsymbol{x}_t}\mathcal L_{\text{CE}}(f(\boldsymbol{x}_{t}), y_{tar}))$ as Eq.~\eqref{eq5};
            \ENDIF
            \STATE $\boldsymbol{x}_{now} = \text{round}(\boldsymbol{x}_{t+1})$;
            \IF {$\|\boldsymbol{x}_{now} - \boldsymbol{x}\|_\infty \geq 1$} 
                \IF {\textbf{untargeted attack and} $f(\boldsymbol{x}_{now})$ != $y_{gt}$}
                    \STATE \textbf{break};
                \ELSIF{\textbf{targeted attack and} $f(\boldsymbol{x}_{now})$ == $y_{tar}$}
                    \STATE \textbf{break};
                \ENDIF
            \ENDIF
        \ENDFOR
        \STATE $\boldsymbol{x}_{adv} = \boldsymbol{x}_{now}$;
        \STATE \textbf{return} $\boldsymbol{x}_{adv}$;
    \end{algorithmic}
\end{algorithm}

Beyond the loose $l_\infty$-norm restriction, existing methods attempt to enhance imperceptibility of attacks by incorporating additional constraints or modules to the optimization process. However, by rethinking the essence of imperceptible attacks, it is evident that if an attack can perform successfully at very small cost, the resulting adversarial examples will naturally possess imperceptibility.
On this basis, we argue that previous complex designs for imperceptibility are actually not necessary. With the proposed dual strategies, namely Dynamic Step Size and Adaptive Early Stop, we unleash the potential of classic PGD, and our PGD-Imperceptible (PGD-Imp) is able to perform essentially imperceptible adversarial attacks.

\subsection{Proposed PGD-Imp: Dynamic Step Size}

\begin{table*}[ht]
    \vspace{-1em}
    \scriptsize
    \caption{Comparisons with other state-of-the-art imperceptible attack methods for untargeted scenario.  We use $\epsilon=8$ and $T=100$ in the proposed PGD-Imp for fair comparison. The running time results are obtained on a same machine with a RTX 3090 GPU.}
    \vspace{-1em}
    \label{tab1}
    \centering
    \begin{tabular}{cllcccccccc}
        \toprule
        {Attacked Models} & Attacks   & Iteration   & Time (s) $\downarrow$  & ASR ($\%$) $\uparrow$ & $l_\infty$ $\downarrow$   & $l_2$ $\downarrow$     & \;PSNR $\uparrow$  & \;SSIM $\uparrow$   & \;FID $\downarrow$   & \;LPIPS $\downarrow$  \\
        \midrule
        \multirow{5}{*}{ResNet-50}  

                                    & AdvDrop \cite{duan2021advdrop}       & 150 & 193       & 96.8      & 0.062     & 3.17      & 41.91     & 0.9872    & 5.57      & 0.0061    \\
                                    & PerC-AL \cite{zhao2020towards}       & 1000 & 4085      & {98.8}      & 0.131     & {2.05}      & {46.35}     & 0.9894    & 8.62      & 0.0029    \\
                                    & SSAH \cite{luo2022frequency}          & 200 & 428       & {99.7}      & 0.033     & 2.65      & 43.73     & {0.9911}    & {4.48}      & {0.0021}  \\

                                    & PGD-Imp (ours)  & 34.2 (avg) & \textbf{57}      & \textbf{100.0}      & \textbf{0.004} (1/255)     & \textbf{0.89}      & \textbf{52.93}     & \textbf{0.9988}    & \textbf{1.12}      & \textbf{0.0003}  \\

        \midrule
        \multirow{5}{*}{VGG-19}  

                                    & AdvDrop \cite{duan2021advdrop}       & 150 & 268    & 97.5      & {0.062}     & {3.23}      & 41.79     & 0.9867   & 5.90    & 0.0061   \\
                                    & PerC-AL \cite{zhao2020towards}       & 1000 & 8671   & \textbf{100.0}     & 0.142     & {2.12}      & {45.92}     & 0.9885   & 10.78   & 0.0028   \\
                                    & SSAH \cite{luo2022frequency}          & 200 & 948    & 85.5      & {0.027}     & 2.35      & 44.62     & {0.9920}   & {4.25}    & {0.0017}   \\

                                    & PGD-Imp (ours)  & 34.2 (avg) & \textbf{109}   & {99.9}      & \textbf{0.004} (1/255)     & \textbf{0.99}      & \textbf{52.07}     & \textbf{0.9983}   & \textbf{1.66}    & \textbf{0.0003}  \\

        \midrule
        \multirow{5}{*}{MobileNet-V2}  

                                    & AdvDrop \cite{duan2021advdrop}       & 150 & 116    & 97.7     & {0.063}     & {3.16}      & {41.94}     & 0.9873   & {4.88}    & 0.0064   \\
                                    & PerC-AL \cite{zhao2020towards}       & 1000 & 3187   & 99.8     & 0.118     & {2.16}      & {45.67}     & 0.9879   & 8.77    & 0.0032   \\
                                    & SSAH \cite{luo2022frequency}          & 200 & 265    & 97.8     & {0.026}     & 2.18      & 45.24     & {0.9930}   & {2.94}    & {0.0016}   \\

                                    & PGD-Imp (ours)  & 67.63 (avg) & \textbf{24}    & \textbf{100.0}     & \textbf{0.004} (1/255)     & \textbf{0.92}      & \textbf{52.59}     & \textbf{0.9986}   & \textbf{1.02}    & \textbf{0.0003}  \\

        \midrule
        \multirow{5}{*}{WideResNet-50}  
                                    
                                    & AdvDrop \cite{duan2021advdrop}       & 150 & 353    & {96.5}     & 0.062     & 3.28      & {41.64}     & {0.9863}   & {6.21}    & {0.0060}  \\
                                    & PerC-AL \cite{zhao2020towards}       & 1000 & 6655   & {97.8}     & {0.133}     & {1.91}      & {46.80}     & {0.9906}   & 9.28    & {0.0025}   \\
                                    & SSAH \cite{luo2022frequency}          & 200 & 738   & 95.7     & {0.028}     & 2.21      & 45.21     & {0.9933}   & {3.95}    & {0.0015}  \\
                                    & PGD-Imp (ours)  & 26.98 (avg) & \textbf{100}   & \textbf{100.0}     & \textbf{0.004} (1/255)     & \textbf{0.90}      & \textbf{52.83}     & \textbf{0.9988}   & \textbf{1.26}    & \textbf{0.0003}  \\

        \bottomrule
    \end{tabular}
    \vspace{-2em}
\end{table*}

To attack at minimal cost, the basic optimization problem of attack can be transformed into finding the adversarial perturbation $\boldsymbol{\delta}$ with minimal strength that allows the adversarial example $\boldsymbol{x}_{adv}=\boldsymbol{x}+\boldsymbol{\delta}$ to cross the decision boundary of attacked model.
Inspired by the well-studied learning rate schedules in finding the optimum for optimization \cite{kingma2014adam, loshchilov2016sgdr, loshchilov2017decoupled, ruder2016overview}, we first introduce the Dynamic Step Size to further refine the iterative process of PGD for a more optimal solution to the transformed optimization problem.

Instead of using a fixed step size to make equal modifications at each step and relying on projection operations to meet the budget $\epsilon$, 
the dynamic step size aims to allocate the total budget to each step unequally for the refinement. 
Specifically, given the iteration steps $T$, the total budget $\|\cdot\|_\infty \leq \epsilon$, and the $\text{sign}(\cdot)$ operation that makes the gradient direction to values of $1$ or $-1$, the dynamic step size $\alpha_t$ at step $t$ should satisfy:
\begin{equation}
\sum_{t=1}^T \alpha_t = \sum_{t=1}^T \eta_t \cdot \beta = \epsilon,
\label{eq3}
\end{equation}
where $\alpha_t = \eta_t \cdot \beta$, $\eta_{1:T} \in {(0, 1]}^T$ is a pre-defined coefficient sequence obtained with a specific schedule (e.g., linear, cosine, etc.), and $\beta$ is a fixed scaling factor. With determined $\eta_t$, we can calculate  $\beta = {\epsilon} / {\sum_{t=1}^T \eta_t}$ and obtain the final dynamic step size $\alpha_t$ for each step $t$.
Thus, the refined process of the proposed PGD-Imp with dynamic step size is expressed as:
\begin{equation}
\boldsymbol{x}_{t+1} = \boldsymbol{x}_t + \eta_t\cdot\beta\cdot\text{sign}(\nabla_{\boldsymbol{x}_t}\mathcal L_{\text{CE}}(f(\boldsymbol{x}_{t}), y_{gt}))\text{, and}
\label{eq4}
\end{equation}
\begin{equation}
\boldsymbol{x}_{t+1} = \boldsymbol{x}_t - \eta_t\cdot\beta\cdot\text{sign}(\nabla_{\boldsymbol{x}_t}\mathcal L_{\text{CE}}(f(\boldsymbol{x}_{t}), y_{tar}))
\label{eq5}
\end{equation}
for untargeted and targeted attacks, respectively.

\subsection{Proposed PGD-Imp: Adaptive Early Stop}
Except for searching the optimal solution, attacking at minimal cost also implies that the final adversarial examples should just barely cross the decision boundary and remain close to it, thus to minimize redundant perturbation strength as much as possible. To this end, we propose Adaptive Early Stop strategy to further reduce the redundant cost and enhance the robustness of attack.

Intuitively, consistent with early stopping to avoid the overfitting during the optimization of training networks\cite{prechelt2002early, liu2008optimized, bai2021understanding}, we expect the optimized adversarial examples not to overfit but just cross the decision boundary of the attacked model.
Thus, the Adaptive Early Stop strategy assesses whether the current result has successfully attacked the classifier at each step in the later iterations in which the accumulated adversarial perturbation can pass through the rounding operation for saving 8-bit images. If successful, the process halts immediately.
Working synergistically with the dynamic step sizes, the attack with Adaptive Early Stop starts with the refined optimization process in the early iterations to confirm the direction of optimization, then progresses
along this direction and finally stops near the decision boundary as soon as current adversarial example misleads the classifier successfully, thereby ensuring a more optimal solution with lower cost and minimizing redundant modifications simultaneously for the essential imperceptibility. Moreover, this strategy also reduces the overall computational complexity and eliminates the sensitivity of PGD-Imp to the hyperparameters $\epsilon$ and $T$.

Algorithm~\ref{alg1} provides the pseudo-code for PGD-Imp. In implementation, we further enhance the computational efficiency and improve GPU utilization by batchifying the calculations and making the early stopped samples no longer participate in subsequent optimization process of the same batch.

\section{Experiments}

\subsection{Experimental Setup}
For untargeted attacks, the experiments are conducted on the widely adopted ImageNet-compatible dataset from the NIPS 2017 adversarial competition  \cite{kurakin2018adversarial} containing 1,000 images of ImageNet \cite{krizhevsky2012imagenet} classes. We include three state-of-the-art imperceptible attack methods, AdvDrop \cite{madry2018towards}, Perc-AL \cite{madry2018towards}, and SSAH \cite{madry2018towards} in our comparison. Four popular image classification backbones ResNet-50 \cite{he2016deep}, VGG-19 \cite{simonyan2014very}, MoblieNet-V2 \cite{sandler2018mobilenetv2}, and WideResNet-50 \cite{zagoruyko2016wide} are used as the attacked models.
For targeted attacks, following AdvINN \cite{chen2023imperceptible}, we use the same 1000 images provided by them and the least-likely objective setting to align with it. 
The attack efficacy is evaluated using Attack Success Rate (ASR), and the imperceptibility is comprehensively assessed using $l_\infty$ and $l_2$ distances,  Peak-Signal-to-Noise Ratio (PSNR), Structure Similarity (SSIM), two learning-based metrics FID \cite{heusel2017gans} and LPIPS \cite{zhang2018unreasonable}.
The compared attack methods use their open-sourced codes and default configurations, and the proposed PGD-Imp uses a linear schedule for $\eta$ and typical settings of $\epsilon=8$, $T=100$.

\begin{table}[t]
    \scriptsize
    \centering
    \caption{Comparisons with other state-of-the-art imperceptible attack methods for targeted attacks against ResNet-50.}
    \vspace{-1em}
    \begin{tabular}{lccccc}
        \toprule
        Attacks   & ASR ($\%$) $\uparrow$ & $l_\infty$ $\downarrow$   & $l_2$ $\downarrow$  & \;SSIM $\uparrow$  & \;LPIPS $\downarrow$  \\
        \midrule
        AdvDrop \cite{duan2021advdrop}       & \textbf{100.0}      & 0.07     & 18.47      & 0.977    & 0.0639    \\
        PerC-AL \cite{zhao2020towards}       & \textbf{100.0}      & 0.10     & {1.93}      & 0.995    & 0.0339    \\
        SSAH \cite{luo2022frequency}          & {99.8}      & 0.03     & 6.97      & {0.991}    & {0.0352}  \\
        AdvINN-CGT \cite{chen2023imperceptible}    & \textbf{100.0}      & 0.03     & 2.66      & \textbf{0.996}    & {0.0118}  \\
        PGD-Imp (ours)  & \textbf{100.0}      & \textbf{0.01}     & \textbf{1.42}      & \textbf{0.996}    & \textbf{0.0009}  \\
        \bottomrule
    \end{tabular}
    \label{tab2}
\end{table}

\vspace{-0.1cm}

\subsection{Comparison with State-of-the-arts}
\textbf{Untargeted attacks.} The comparison results are shown in Table~\ref{tab1}. By rethinking the essence of imperceptible adversarial attacks, the proposed PGD-Imp method consistently achieves the best results in both ASR and imperceptibility simultaneously, significantly outperforming previous methods. For all the victim models, our PGD-Imp achieves an $l_2$ distance that is about only one-quarter to one-third that of other methods, and $l_\infty$=0.004 (1/255) indicates that PGD-Imp only modifies each pixel of the image by a maximum of $\pm 1$, demonstrating its ability to push the image to the decision boundary of the attacked model with minimal cost. Compared to SSAH, when attacking ResNet-50, our PGD-Imp achieves 100$\%$ ASR with tremendous improvements of 52.93 (+9.2) PSNR and 0.9988 (+0.0077) SSIM, 1.12 (-3.36) FID, and 0.0003 (-0.0018) LPIPS for image quality. Moreover, PGD-Imp also maintains a leading edge in terms of running time and number of iterations thanks to the two proposed cooperating strategies.

\textbf{Targeted attacks.} 
As shown in Table~\ref{tab2}, the proposed PGD-Imp can also perform more challenging targeted attacks. Our PGD-Imp continues to outperform the advanced AdvINN-CGT that is specifically designed for targeted attacks with an additional invertible neural network, achieving 100$\%$ ASR but with only half the attack cost of 0.01 (-67$\%$) $l_\infty$, 1.42 (-47$\%$) $l_2$ distances and better image quality of 0.0009 LPIPS.

\textbf{Visualization.} As shown in Fig.~\ref{fig2}, it is evident that even after amplification of perturbations and zooming into specific regions, there is still almost no noise pattern can be detected in the adversarial example crafted by the proposed PGD-Imp whereas all other methods show some artifacts, and the overall perturbation intensity of PGD-Imp is also obviously lower.

\begin{table}[t]
    \scriptsize
    \centering
    \caption{From PGD to PGD-Imp: ablation study on the proposed Dynamic Step Size (DSS) and Adaptive Early Stop (AES) strategies.}
    \vspace{-1em}
    \setlength{\tabcolsep}{1.5pt}
    \begin{tabular}{lcccccc}
        \toprule
        Attacks  & Value of $\epsilon$  & Iter.  & ASR ($\%$) $\uparrow$   & $l_2$ $\downarrow$  & \;PSNR $\uparrow$  & \;SSIM $\uparrow$  \\
        \midrule
        PGD         & $\epsilon=2$ & 10  & \textbf{100.0}     & 2.53       & 43.72    & 0.9853   \\
        PGD         & $\epsilon=2$ & 100  & \textbf{100.0}     & 2.44       & 44.03    & 0.9873   \\
        PGD + DSS   & $\epsilon=2$ & 100  & {99.9}     & 1.75       & 46.99    & 0.9952   \\
        PGD + AES   & $\epsilon=2$ & 9.9 (100) & \textbf{100.0}     & 1.60       & 47.77    & 0.9937   \\
        PGD-Imp (DSS $\&$ AES)     & $\epsilon=2$ & 56.3 (100)  & {99.9}     & \textbf{0.86}     & \textbf{53.29}  & \textbf{0.9990} \\
        \bottomrule
    \end{tabular}
    \label{tab3}
    \vspace{-0.4cm}
\end{table}

\begin{table}[t]
    \scriptsize
    \centering
    \caption{Ablation study on the schedules of $\eta_t$ in Dynamic Step Size with fixed $\epsilon=8$ and $T=100$ against ResNet-50.}
    \vspace{-1em}
    \setlength{\tabcolsep}{3pt}
    \begin{tabular}{lcccccc}
        \toprule
        Schedules & Trend of $\eta_t$  & Iter.  & ASR ($\%$) $\uparrow$   & $l_2$ $\downarrow$  & \;PSNR $\uparrow$  & \;SSIM $\uparrow$  \\
        \midrule
        Constant       & $\eta_t\equiv1$          & 31.9  & \textbf{100.0}     & 0.96       & 52.30  & 0.9986   \\
        Cosine-reverse \ \  & $1 \rightarrow 0$     & 34.4 & \textbf{100.0}     & 1.03       & 51.59  & 0.9982   \\
        Cosine         & $0 \rightarrow 1$            & 42.0 & \textbf{100.0}     & 0.90       & \textbf{52.93}  & \textbf{0.9988}   \\
        Linear-reverse \ \ & $1 \rightarrow 0$    & 34.4 & \textbf{100.0}     & 1.03       & 51.57  & 0.9981 \\
        Linear (PGD-Imp)        & $0 \rightarrow 1$   & 34.2 & \textbf{100.0}     & \textbf{0.89}       & \textbf{52.93}  & \textbf{0.9988} \\
        \bottomrule
    \end{tabular}
    \label{tab4}
\end{table}

\subsection{Ablation Study}
As shown in Table~\ref{tab3}, we first conduct ablation study on transitioning from the PGD with a lower budget ($\epsilon=2$) to the proposed PGD-Imp against ResNet-50 to verify the effectiveness of the proposed DSS and AES strategies. 
Compared to the original PGD, the results in the third row indicate that DSS helps the adversarial example converge to a more optimal solution with lower attack cost (i.e., $l_2$ from 2.44 to 1.75) under the same number of iterations. Meanwhile, the fourth row shows that AES also can reduce redundant perturbations in the adversarial example (i.e., $l_2$ from 2.44 to 1.60), allowing it to be located just near the decision boundary of the attacked model. Ultimately, the final PGD-Imp achieves a more significant improvement benefiting from the cooperation and mutually reinforcing effects of the proposed DSS and AES strategies.

Furthermore, we also study the schedule for determining  $\alpha_t=\eta_t\cdot\beta$ in the DSS strategy. Table~\ref{tab4} shows that using a linear schedule yields the best performance. We suppose this is because the gradually increasing step size starting from 0 aligns with the ideal condition of PGD-Imp with DSS and AES. The refined optimization process with DSS explores and progresses in a direction toward an optimal solution for reducing the attack cost in the early iterations, and this process stops immediately if attacking successfully to further minimize the redundant perturbations using AES. In this way, step sizes increasing from 0 to 1 are even more finer than those from 1 to 0 before the stopping, and the step size at the stopping iteration is also smaller, making the two strategies more effective.

\begin{figure}[t]
\centering
\includegraphics[width=0.85\columnwidth]{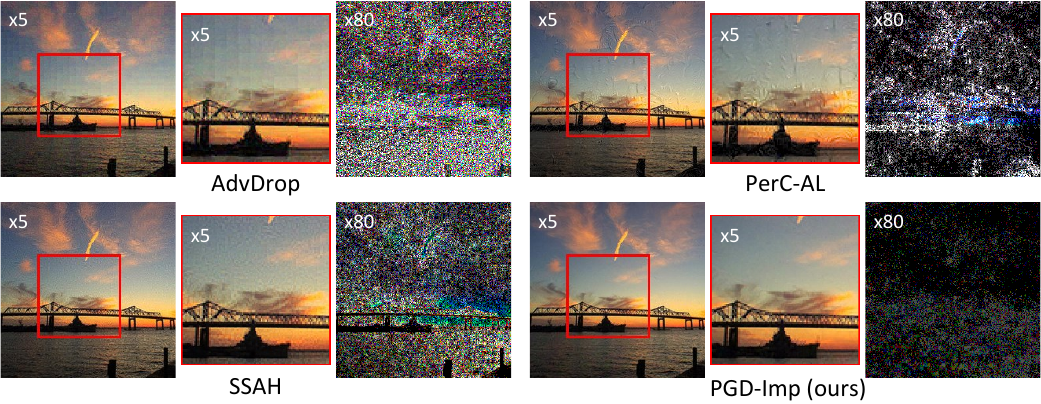}
\vspace{-0.8em}
\caption{Visualization of four imperceptible attacks for untargeted scenario. Perturbations within the images are amplified as marked for better observation.
}
\label{fig2}
\vspace{-0.25cm}
\end{figure}

\begin{figure}[t]
\centering
\includegraphics[width=0.88\columnwidth]{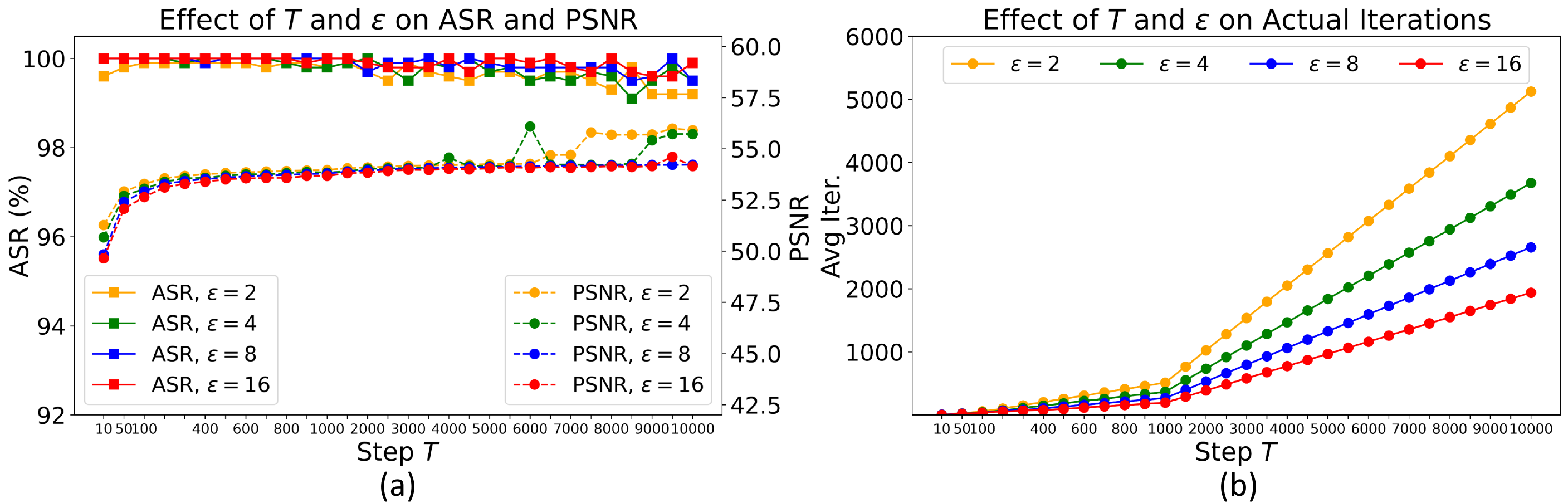}
\vspace{-1em}
\caption{Experimental results on the effect of Step $T$ and $\epsilon$ on the proposed PGD-Imp. The attacked model is ResNet-50.}
\label{fig3}
\end{figure}

\subsection{Discussion}
To explore the working mechanism and extreme performance of PGD-Imp, we conducted an in-depth discussion by using more values of the only two hyperparameters $T$ and $\epsilon$. The experimental results are shown in Fig.~\ref{fig3}.
Firstly, in PGD-Imp, larger $T$ and smaller $\epsilon$ correspond to a more refined optimization process. In other words, the imperceptibility of the attack should be proportional to $T$ and inversely proportional to $\epsilon$, as reflected in the results of Fig.~\ref{fig3}(a). For the extreme cases, the performance of PGD-Imp is fully squeezed with a larger $T \geq 8000$ and smaller $\epsilon=2$ or $\epsilon=4$, achieving an impressive PSNR $\approx 56$ with ASR $\approx 100 \%$.

Secondly, the collaboration between the proposed dual strategies also enhances the robustness of PGD-Imp, making it less sensitive to these two hyperparameters. In the most cases of Fig.~\ref{fig3}(a), regardless of how $T$ and $\epsilon$ change, it  stably remains PSNR $\geq53$ and ASR $\geq99\%$. Only when $T=10$, there is a slight performance drop due to the insufficient number of iterations to reach the optimization solution.

Finally, as shown in Fig.~\ref{fig3}(b), since $\eta_t$ adopts a linear schedule as mentioned above, the actual number of required iterations is linearly related to $T$ when $\epsilon$ is fixed. Moreover, for the same $T$, the larger the epsilon, the fewer iterations are actually required, corresponding to a shorter running time. Because as $\epsilon$ increases, each iteration is allocated a larger step size, allowing the AES to be triggered at an earlier stage.

\section{Conclusion}
In this paper, we rethink imperceptible adversarial attacks and realize that the essence does not lie in additional constraints but rather in pushing the adversarial example across the decision boundary of the attacked model with minimal attack cost. Based on this insight, we propose PGD-Imp, a novel imperceptible attack with two simple yet highly effective strategies from an optimization perspective for the first time. Extensive experiments demonstrate that PGD-Imp achieves state-of-the-art performance in both untargeted and targeted imperceptible attacks.
For future work, the proposed PGD-Imp is possibly to be combined with more advanced designs or promote corresponding defenses like using the classic PGD, and it also holds the potential to determine the model decision boundaries and benefit the research on model interpretability.


\bibliographystyle{ieeetr}
\bibliography{references}

\end{document}